\documentclass[conference]{IEEEtran}
\usepackage{cite}
\usepackage{amsmath,amssymb,amsfonts}
\usepackage{algorithmic}
\usepackage{graphicx}
\usepackage{textcomp}
\usepackage{xcolor}
\usepackage{hyperref}

\usepackage{caption}
\usepackage{subcaption}
\usepackage{makecell}
\usepackage{multirow}
\usepackage{placeins}

\begin{document}

\title{SOLAQUA: SINTEF Ocean Large Aquaculture Robotics Dataset}

\author{\IEEEauthorblockN{Sveinung Johan Ohrem$^{1,\dagger}$, Bent Haugaløkken$^1$, Eleni Kelasidi$^{1,2}$}
\IEEEauthorblockA{$^1$ Department of Aquaculture, SINTEF Ocean, Trondheim, Norway\\
$^2$ Department of Mechanical and Industrial Engineering, NTNU, Trondeim, Norway\\
$^{\dagger}$ Corresponding author: sveinung.ohrem@sintef.no}
}

\maketitle

\begin{abstract}
This paper presents a dataset gathered with an underwater robot in a sea-based aquaculture setting. Data was gathered from an operational fish farm and includes data from sensors such as the Waterlinked A50 DVL, the Nortek Nucleus 1000 DVL, Sonardyne Micro Ranger 2 USBL, Sonoptix Mulitbeam Sonar, mono and stereo cameras, and vehicle sensor data such as power usage, IMU, pressure, temperature, and more. Data acquisition is performed during both manual and autonomous traversal of the net pen structure. The collected vision data is of undamaged nets with some fish and marine growth presence, and it is expected that both the research community and the aquaculture industry will benefit greatly from the utilization of the proposed SOLAQUA dataset.
\end{abstract}

\section{Introduction}
Aquaculture is and will be an important contributor to the production of protein and food in the years to come. Aquaculture as an industry is found all over the world, with varying production methods dominating depending on the country~\cite{fore2018precision}. In Norway, sea-based aquaculture is the current industry standard, as the Norwegian geography with sheltered fjords invites this production type. Atlantic salmon (\emph{salmo salar}) is the most commonly farmed species and the production method of choice is net pens. These net pens, and the surrounding infrastructure, are in constant need of inspection, maintenance and repair (IMR) operations. The aquaculture industry in Norway has adapted underwater remotely operated vehicles (ROVs) for majority of the inspection and intervention tasks in fish farms. 

Typical ROV operations in aquaculture range from net and mooring inspections and cleaning, to fish monitoring and net pen installations~\cite{kelasidi2023robotics}. 

A net pen offers a unique domain for underwater robots: The operations are normally performed in - or close to - the splash zone with currents and waves affecting the robot's states. The environment is dynamic, with large moving structures and several potentially moving obstacles. There are hundreds of thousands of fish in a net pen which will interfere sensor signals and block camera views. All of this combined with the risks and costs of performing sea-based operations makes aquaculture a challenging domain for underwater robotics which is likely to benefit greatly from the development of higher levels of autonomy~\cite{kelasidi2023robotics}.

Research efforts have been made over the last years to advance the level of autonomy in the mentioned areas as these operations are mostly manually performed today~\cite{amundsen2024aquaculture}. Early results include novel methods for navigation in net pens using a Doppler velocity logger (DVL) and ultra-short baseline (USBL) system~\cite{rundtop2016experimental}. This work led to the development of a net-following algorithm~\cite{amundsen2021autonomous} which has been demonstrated in several later works, e.g., ~\cite{ohrem2023adaptive,ohrem2024adaptive,haugalokken2023adaptive}. Navigation using low-cost sensor systems was investigated in~\cite{haugalokken2024low} where various low-cost sonar and DVL technologies are shown to be suitable for navigation purposes in net pens. Camera-based methods are investigated, for example, in~\cite{akram2022visual,schellewald2021vision} with the goal of reducing the number of navigation sensors needed on underwater vehicles in aquaculture.

\begin{figure}[ht]
     \centering
     \begin{subfigure}{0.5\columnwidth}
         \centering
         \includegraphics[height=4cm]{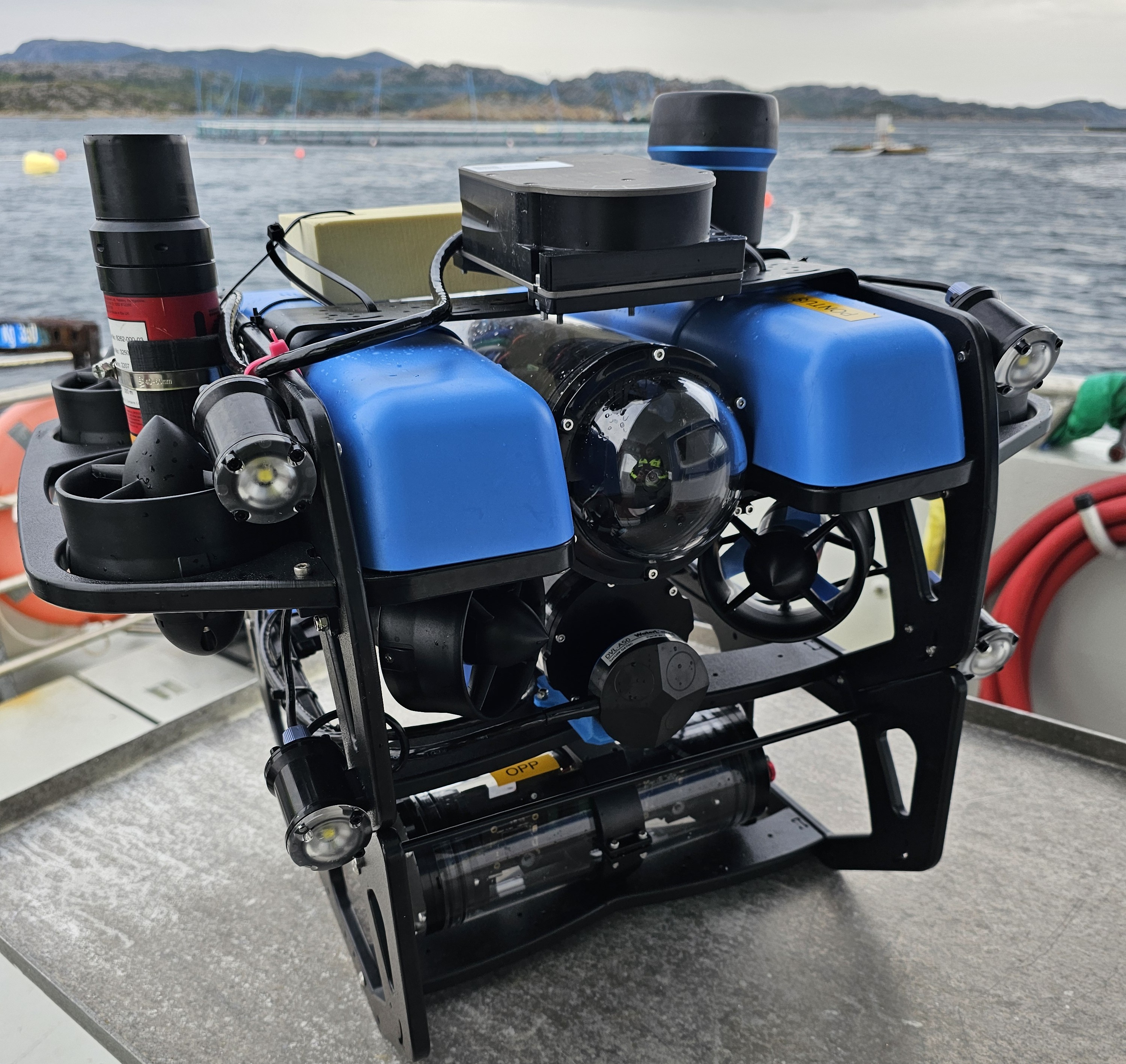}
         \caption{ROV with sensor configuration 1.}
         \label{fig:ROV_config1}
     \end{subfigure}
     \begin{subfigure}{0.4\columnwidth}
         \centering
         \includegraphics[height=4cm]{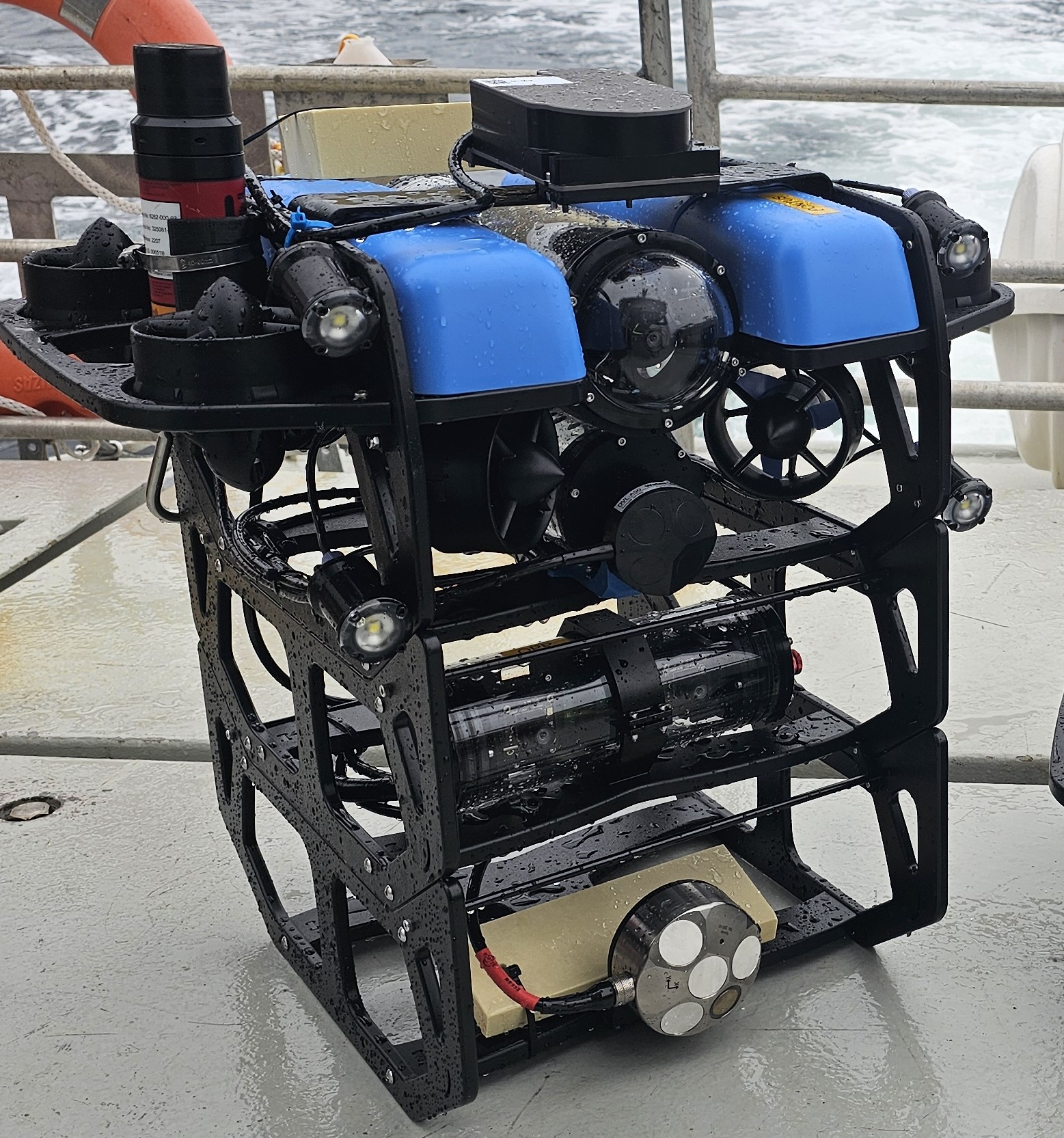}
         \caption{ROV with sensor configuration 2.}
         \label{fig:ROV_config2}
     \end{subfigure}
        \caption{ROV with two different sensor configurations.}
        \label{fig:ROV_configurations}
\end{figure}

As methods driven by artificial intelligence (AI) and machine learning (ML) are adapted in the aquaculture robotics domain, the need for datasets to train AI and ML models is of great importance to contribute and advance the research on robust methods for fully autonomous IMR operations, and thus address current and future challenges of the aquaculture industry. Particularly in the field of vision-based and multi-modal navigation, where camera images paired with e.g., position measurements from other sensors such as DVL and USBL can have significant value. 

To this end, we present the SINTEF Ocean Large Aquaculture Dataset, SOLAQUA, consisting of navigation sensor data, sonar images, mono and stereo camera images, and vehicle data, all captured in a fully operational fish farm. The dataset was captured while performing manual control of a remotely operated vehicle (BlueROV2) as well as during autonomous traversal of the net structure using the previously developed net following algorithm. The data may be utilized for the development of navigation methods, developing novel computer vision techniques for navigation and inspection, simulation purposes, to verify novel control methods, and much more. Examples of use of the dataset for development of robust localization and mapping methods for UUVs operating in industrial scale fish farms can be found in e.g.,~\cite{botta2024framework}.

The dataset consists of numerous subsets of data from an ROV operating between 1 and 2 minutes at different depths, with various speeds, net-relative distances, and facing different environmental conditions. The ROV faces the net structure during the runs, hence images are of undamaged nets with some marine growth. Some fish are present, but the purpose of this dataset is not to capture images of fish. To our knowledge, no similar dataset exist.

This paper briefly describes the platform used for gathering the data (a BlueROV2) and the data acquisition procedure. In addition, information are presented to give detailed overview of the sensors, their logging frequency and position on the ROV. This overview together with the concrete examples of data found in the dataset can be utilized from the research community to develop and propose novel methods.

The dataset can be found at the following website: \url{https://data.sintef.no/feature/fe-a8f86232-5107-495e-a3dd-a86460eebef6} and is under CC BY-SA license.

\section{Multi-modal underwater platform for data collection and autonomous operations}
In the presented dataset the vehicle was either moved manually or autonomously using net following. Net following refers to an automatic control procedure where the vehicle moves relative to the net at a desired distance, heading angle, and velocity relative to the net structure~\cite{amundsen2021autonomous}. All datasets are logged with time-synchronization using ROS (Robot Operating System), i.e., the time-stamp of each logged data point is generated by the same clock. Due to varying data capturing frequencies between the sensors and generated signals, minor variations in time-stamps, in the millisecond range, are present. No known delays occurred while capturing the data. The data was logged as .bag-files.

As shown in Table~\ref{tab:rov_specs} and Figure~\ref{fig:ROV_configurations} two configurations of the BlueROV2 was used to gather the datasets. Each configuration results in a vehicle with different dimensions and weights, and with a variety of integrated relevant sensors. An overview of the ROV system architecture with detailed diagrams illustrating the connections between sensors, actuators, control system and communication interface can be found in~\cite{haugalokken2024low}.

\begin{table}
\caption{BlueROV2 Technical Specifications}
\centering
\begin{tabular}{l|l}
\hline
\textbf{Feature} & \textbf{Description} \\
\hline
Size (L x W x H) (Config 1) & 457 x 436 x 397 [mm] \\

Size (L x W x H) (Config 2) & 457 x 436 x 541 [mm] \\

Weight in air/water (Config 1) & 15/0 [kg] \\

Weight in air/water (Config 2) & 18/0 [kg] \\

Number of thrusters & 8 \\

Communication & Ethernet, UDP \\

Degrees of Freedom & 6 \\

Power & Battery, 14.8 [V] \\
\hline
\end{tabular}
\label{tab:rov_specs}
\end{table}

\subsection{Sensors}
A list of sensors used to gather the datasets is shown in Table~\ref{tab:sensors}. Note that some custom ROS messages have been used. These can be downloaded alongside the respective datasets. Frequency values may sometimes vary slightly throughout a test (denoted with a $\sim$ symbol). For acoustic sensors, the frequency may vary depending on the distance between the sensor and the target object. For camera data, datasets contain either {\fontfamily{qcr}\selectfont
Image} or {\fontfamily{qcr}\selectfont
CompressedImage} (ROS) message types, where more recent trials have chosen to use {\fontfamily{qcr}\selectfont
CompressedImage} to reduce data size.

\begin{table}[h!]
\caption{Overview of performed tests}
    \centering
    \begin{tabular}{l|l} \hline
       \textbf{Purpose of test}  & \textbf{Number of sets} \\ \hline
        Camera Calibration (CC) & 3 \\ 
        Manual Control (MC) & 6 \\ 
        Net following horizontal (NFH) & 54 \\ \hline
    \end{tabular}
    \label{tab:test_overview}
\end{table}

\begin{table}
\caption{Mono camera parameters}
\centering
\begin{tabular}{|l|l|}
\hline
Camera matrix & 
\(\begin{bmatrix}
348.5190 & 0 & 317.4431 \\
0 & 349.9505 & 168.9469 \\
0 & 0 & 1.0000 \\
\end{bmatrix}\) \\ \hline
Reprojection error & 0.191835 \\ \hline
Distortion coefficients & \([0.0117, 0.0300, -0.0328, 0, 0]\) \\ \hline
\end{tabular}
\label{tab:mono_camera_params}
\end{table}

\begin{table*}[htbp]
\caption{Sensor Specifications and ROS Topics}
\centering
\renewcommand{\arraystretch}{1.3}
\begin{tabular}{|l|l|l|l|l|}
\hline
\textbf{Sensor} & \textbf{Frequency} & \textbf{Additional info} & \textbf{ROS topic} & \textbf{ROS Message name} \\
\hline
IMU & \makecell[l]{$\sim15-25$Hz\\ $\sim15-25$Hz }& \makecell[l]{Acc, Gyro \\ $\Theta$ (includes the IMU's internal \\ estimates of roll, pitch and yaw)} & \makecell[l]{/sensor/imu \\ /sensor/attitude*} & \makecell[l]{ IMU \\ Attitude} \\
\hline
Barometer & $<$10 Hz & Pressure, depth, temperature & /sensor/depth\_temperature & DepthTemperature \\
\hline
Ping Echosounder & $\sim10$Hz & \makecell[l]{Includes distance and \\ confidence on measurement} & /sensor/ping & Ping \\
\hline
Ping360 & $<$25 Hz & \makecell[l]{Logged ``range'' data is \\ experimental and may be \\inaccurate in trials from \\ 2023-2024} & /sensor/ping360 & Ping360 \\
\hline
DVL DVL-A50 & $\sim10$ Hz & - & \makecell[l]{/sensor/dvl\_position, \\ /sensor/dvl\_velocity} & \makecell[l]{DVLPosition \\ DVLBeam, DVLVelocity} \\
\hline
\multirow{4}{*}{DVL Nucleus1000} & 1-8 Hz & - & /nucleus1000dvl/bottomtrack & Nucleus1000\_bottomtrack \\
 & 100 Hz & - & /nucleus1000dvl/imu & Nucleus1000\_imu \\
 & 10 Hz & - & /nucleus1000dvl/ins & Nucleus1000\_ins \\
 & 50 Hz & - & /nucleus1000dvl/magnetometer & Nucleus1000\_magnetometer \\
\hline
USBL: MicroRanger 2 & 0.5-2 Hz & - & /sensor/usbl & SonardyneUSBL2 \\
\hline
Multi-beam sonar: Sonoptix Echo & $<$25 Hz & - & /sensor/sonoptix\_echo/image & SonoptixECHO \\
\hline
Camera & 15/25 FPS & RGB, 1080p / 720p & \makecell[l]{/bluerov2/image \\ /image/compressed\_image/data} & \makecell[l]{sensor\_msgs/Image \\ sensor\_msgs/CompressedImage} \\
\hline
Stereo camera & 15/25 FPS & RGB, 1080p / 720p & - & \makecell[l]{sensor\_msgs/Image \\ sensor\_msgs/CompressedImage \\ sensor\_msgs/Image \\ sensor\_msgs/CompressedImage} \\
\hline
\end{tabular}
\label{tab:sensors}
\end{table*}

\subsection{Sensors and sensor placements for different vehicle configurations}

\subsubsection{Sensor placement}
The sensor placements are given relative to the vehicle's IMU, and is given in the vehicle's BODY-frame. Units are meters and degrees. The electronics enclosure (4"), which houses the main electronics including the camera and the IMU, has a length of 0.3 meters (which can be used in case you wish to use the center of origin (CO) as your origin). The DVL is intentionally facing forwards to measure distance and speed relative to the net structure in front of the ROV. Sensor placements are given in Table~\ref{tab:sensor_placement}.

\begin{table*}
\caption{Sensor Configurations. Units are meters and degrees.}
\centering
\begin{tabular}{|l|l|l|}
\hline
Sensor & Configuration 1 [x, y, z, $\phi$, $\theta$, $\psi$]& Configuration 2 [same as conf. 1] \\ \hline
IMU & [0, 0, 0, 0, 0, 0] & (See Configuration 1) \\ \hline
Barometer & [-0.21, 0, 0, 0, 0, 0] & (See Configuration 1) \\ \hline
Ping Echosounder & [0.05, -0.06, 0.13, 0, 0, 0] & N/A \\ \hline
Ping360 & [0.07, -0.08, -0.16, 0, 0, -180.0] & (See Configuration 1) \\ \hline
DVL DVL-A50 & [0.10, 0.04, 0.13, -90, 0, -90] & (See Configuration 1) \\ \hline
DVL Nucleus1000 & N/A & [0.10, 0.04, 0.11, -90, 0, -90] \\ \hline
USBL & [0.08, 0.24, -0.15, 0, 0, 0] & (See Configuration 1) \\ \hline
Multi-beam sonar & N/A & [0.09, 0.04, -0.11, 0, 5, 0] \\ \hline
Camera & [0.09, 0.04, 0, 0, 0, 0] & (See Configuration 1) \\ \hline
Stereo camera (L) & [0.04, 0.095, 0.26, 0, 0, 0] & (See Configuration 1) \\ \hline
Stereo camera (R) & [0.04, -0.015, 0.26, 0, 0, 0] & (See Configuration 1) \\ \hline
\end{tabular}
\label{tab:sensor_placement}
\end{table*}

\begin{table*}
\caption{Stereo camera parameters}
\centering
\begin{tabular}{|l|l|}
\hline
Translation (baseline) & \([-110.16, 0.14, 0.65]\) \\ \hline
Mean Reprojection Error & 0.4170 \\ \hline
Fundamental matrix & 
\(\begin{bmatrix}
-0.0000 & 0.0000 & -0.0016 \\
-0.0000 & -0.0000 & 0.1363 \\
0.0004 & -0.1364 & 0.5083 \\
\end{bmatrix}\) \\ \hline
Essential matrix & 
\(\begin{bmatrix}
-0.0009 & 3.4838 & 0.0517 \\
-0.8839 & -1.2784 & 110.2245 \\
0.0252 & -110.1729 & -1.2772 \\
\end{bmatrix}\) \\ \hline
CAMERA1 Focal Length & \([884.1997, 823.9808]\) \\ \hline
CAMERA1 Principal Point & \([637.8639, 354.3048]\) \\ \hline
CAMERA1 Radial Distortion & \([0.2474, 0.3499]\) \\ \hline
CAMERA1 Tangential Distortion & \([0, 0]\) \\ \hline
CAMERA1 Skew & 0 \\ \hline
CAMERA1 Camera matrix & 
\(\begin{bmatrix}
884.1997 & 0 & 637.8639 \\
0 & 823.9808 & 354.3048 \\
0 & 0 & 1.0000 \\
\end{bmatrix}\) \\ \hline
CAMERA2 Focal Length & \([877.3510, 817.2770]\) \\ \hline
CAMERA2 Principal Point & \([691.8459, 342.6457]\) \\ \hline
CAMERA2 Radial Distortion & \([0.2532, 0.2153]\) \\ \hline
CAMERA2 Tangential Distortion & \([0, 0]\) \\ \hline
CAMERA2 Skew & 0 \\ \hline
CAMERA2 Camera matrix &
\(\begin{bmatrix}
877.3510 & 0 & 691.8459 \\
0 & 817.2770 & 342.6457 \\
0 & 0 & 1.0000 \\
\end{bmatrix}\) \\ \hline
\end{tabular}
\label{tab:stereo_camera_params}
\end{table*}

\subsection{Camera calibration parameters}
Matlab (R2022b) and the in-built camera calibration apps {\fontfamily{qcr}\selectfont
Camera Calibrator} and {\fontfamily{qcr}\selectfont Stereo Camera Calibrator} have been used to obtain the camera parameters (intrinsic, extrinsic). In the calibration procedure, images were collected by recording a video containing a checkerboard at various locations within the video frames. The procedure for performing the calibration can be found in the documentation of the camera calibration apps. The parameters for the mono and stereo camera systems can be found in Table~\ref{tab:mono_camera_params} and Table~\ref{tab:stereo_camera_params}, respectively. Distances are given in millimeters. 

In general, obtaining absolute ground truth measurements in underwater environments is challenging. To address this challenge and obtain ground truth distances to the net, passive visual markers (i.e. AprilTags) have been placed directly on the net surface during field trials, see Figure.~\ref{fig:mono_camera_april_tag}.

\section{Datasets}
Datasets contain tests that may be associated with two log-files, one for video/images (including sonar data), and one for all other data, denoted "timestamp"\_video.bag and "timestamp"\_data.bag, respectively. 

\subsection{Data gathering procedures and methods}
The data was gathered using a BlueROV2 vehicle. Each run was performed using either manual control or autonomous net following. Some runs were also performed solely to gather data for stereo camera calibration. The path driven during the runs was more or less the same for every run, but the depth, speed and distance to the net may vary both during the runs and between runs. See details on the dataset website.

In the dataset you may find the following abbreviations: Manual Control (MC), Net-following (NF), Net-following horizontal (NFH), and Camera Calibration (CC). The dataset was recorded in August 2024. The number of datasets for each performed test are presented in Table~\ref{tab:test_overview}.
\subsubsection{Manual control}
Logging of data started when the vehicle was at the desired depth and distance to the net. The pilot used the \emph{depth hold} mode of the BlueROV2, i.e., the depth and the angles (roll, pitch, yaw) are automatically controlled. The pilot attempted to perform a net following maneuver, i.e., maintaining a certain distance to the net while moving sideways.

\subsubsection{Autonomous net following}
Logging of data started when the vehicle was at the desired depth and at a certain distance from the net not too far from the initial desired distance. The net following was then performed, i.e., the vehicle was automatically commanded to move sideways with a certain speed, distance and heading relative to the net. The speed of the ROV was controlled using the adaptive controller from~\cite{ohrem2024application}. The depth was controlled by a decoupled depth controller.

\section{Example data}
\begin{figure}
    \centering
    \includegraphics[width=\columnwidth]{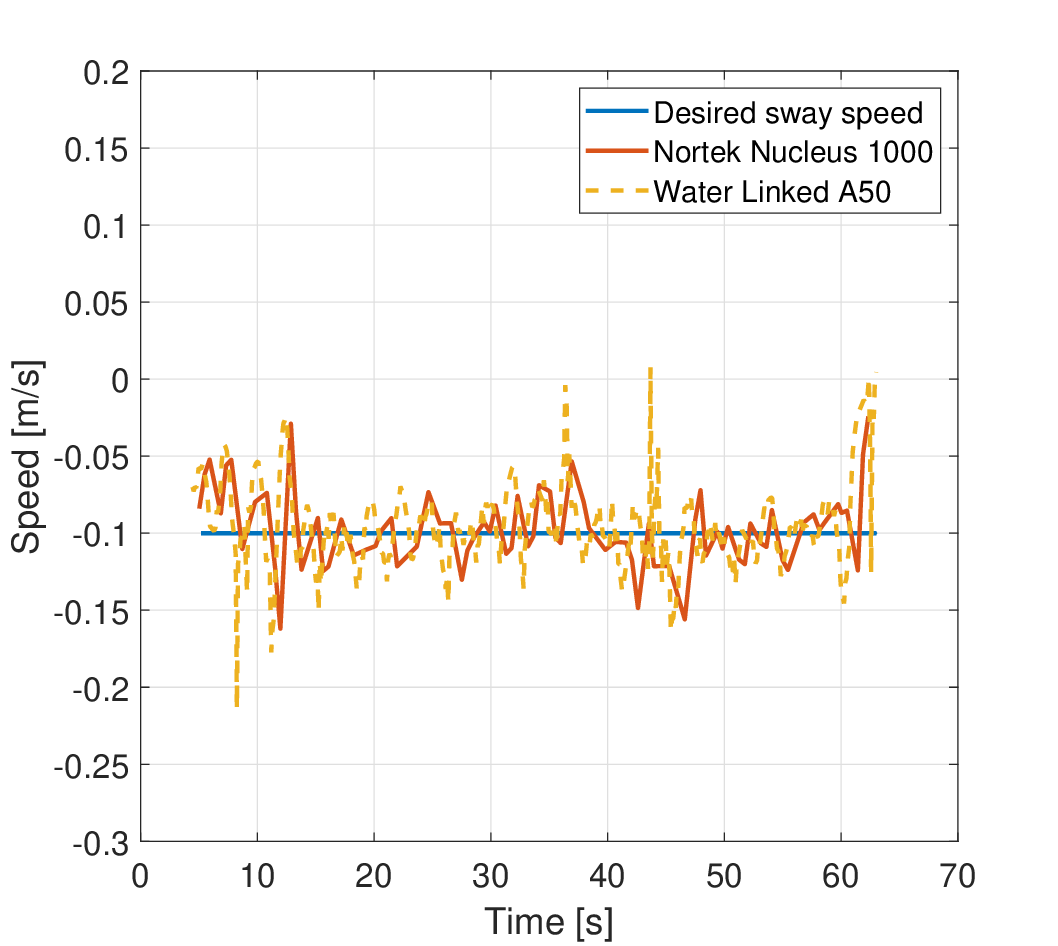}
    \caption{The net-relative sway speed measured by the Water Linked A50 and the Nortek Nucleus 1000 DVL.}
    \label{fig:both_DVLs}
\end{figure}

\begin{figure}
    \centering
    \includegraphics[width=\columnwidth]{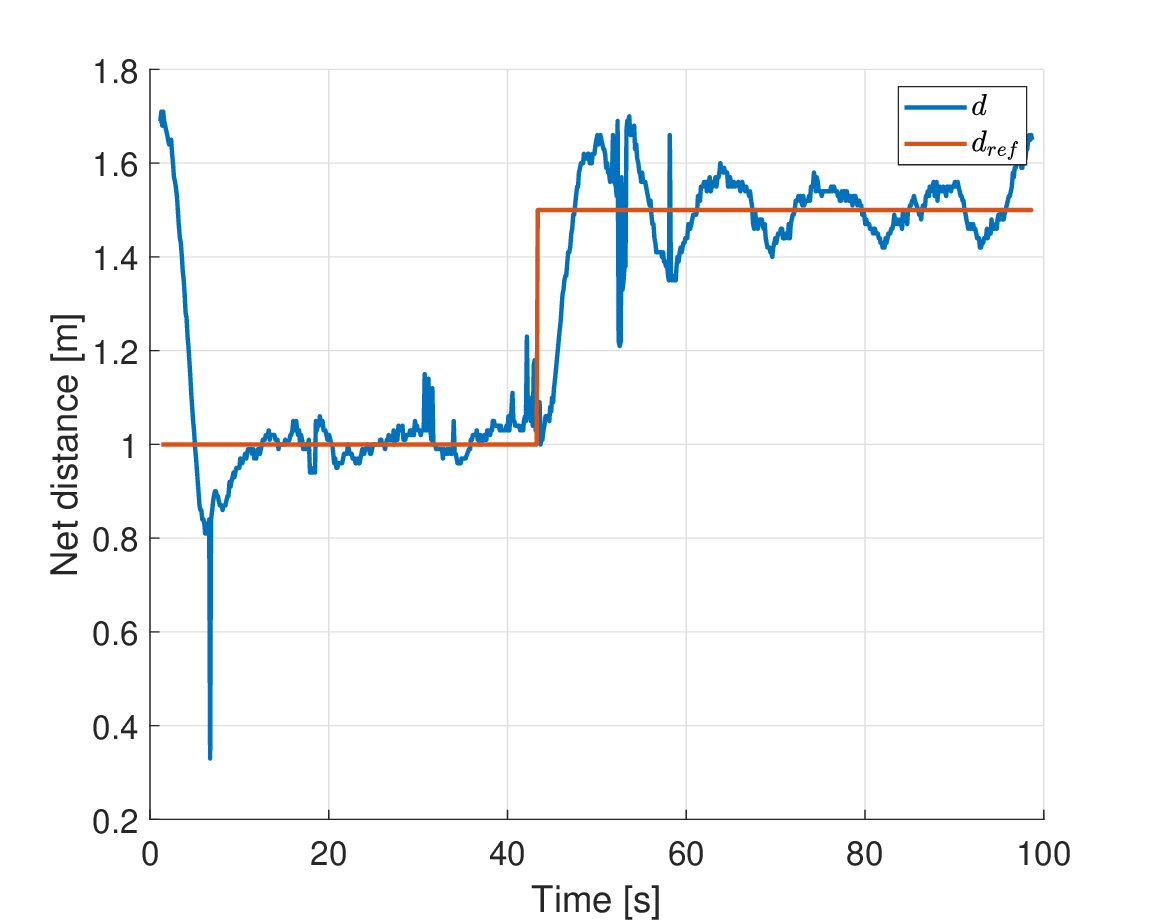}
    \caption{ROV distance to net (blue) and desired distance (red).}
    \label{fig:net_distance}
\end{figure}

\begin{figure}
    \centering
    \includegraphics[width=\columnwidth]{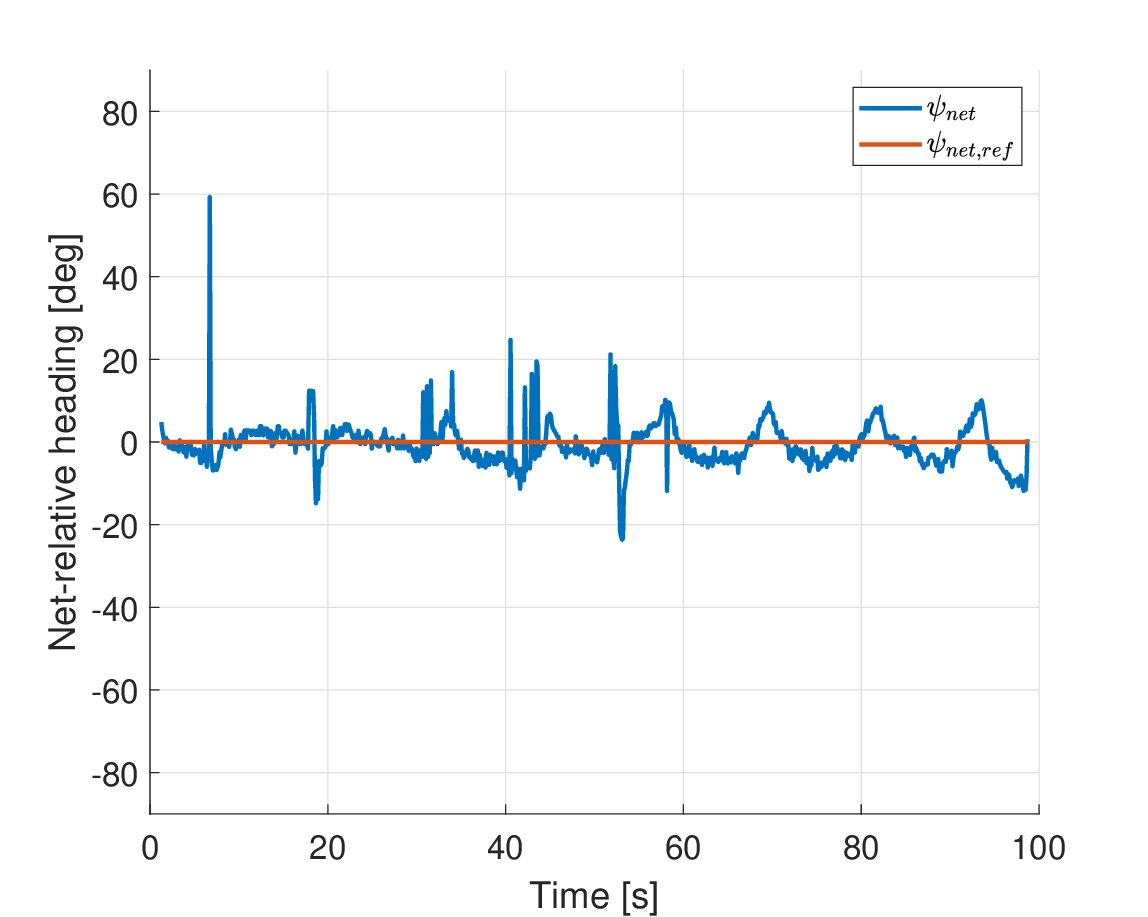}
    \caption{ROV heading relative to net (blue) and desired net relative heading (red).}
    \label{fig:net_heading}
\end{figure}

\begin{figure}
    \centering
    \includegraphics[width=\columnwidth]{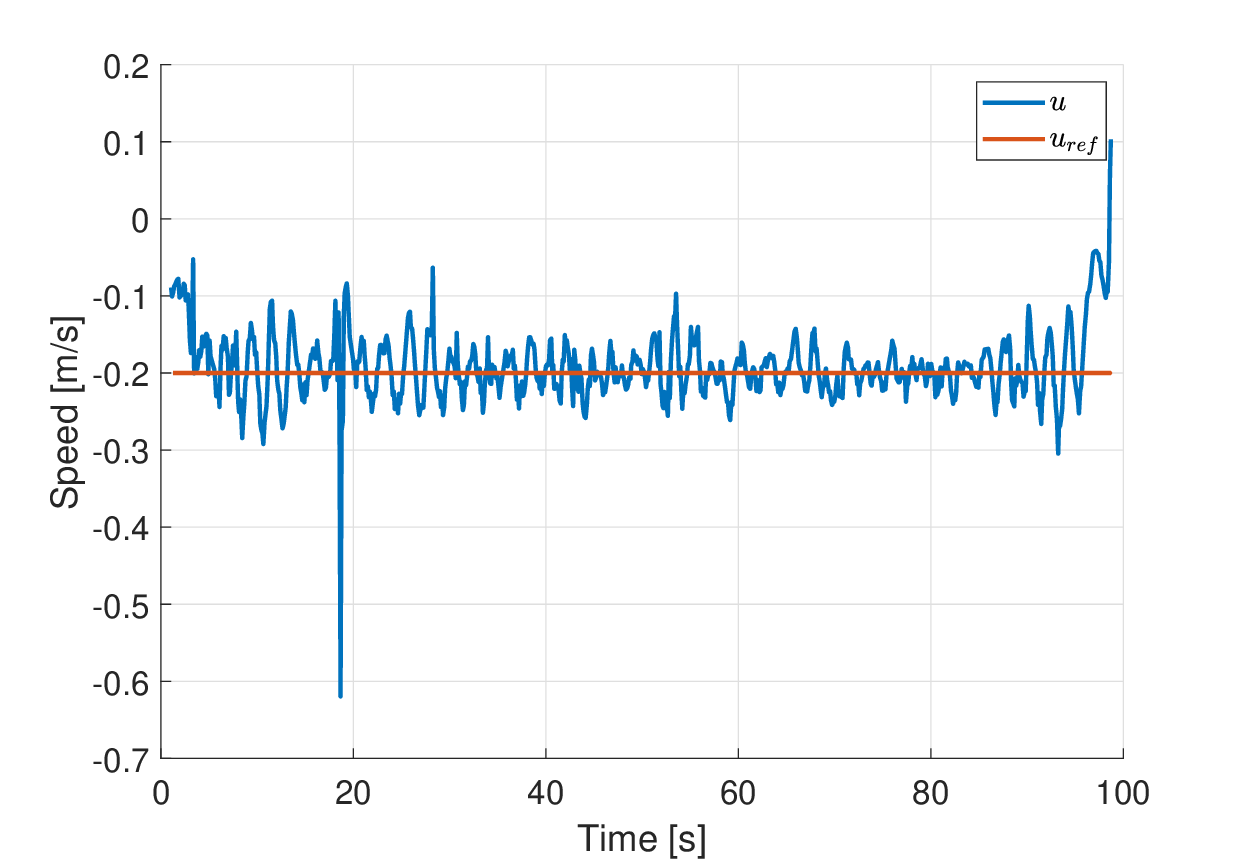}
    \caption{ROV net relative sway speed (blue) and desired speed (red).}
    \label{fig:net_velocity}
\end{figure}

\begin{figure}
    \centering
    \includegraphics[width=\columnwidth]{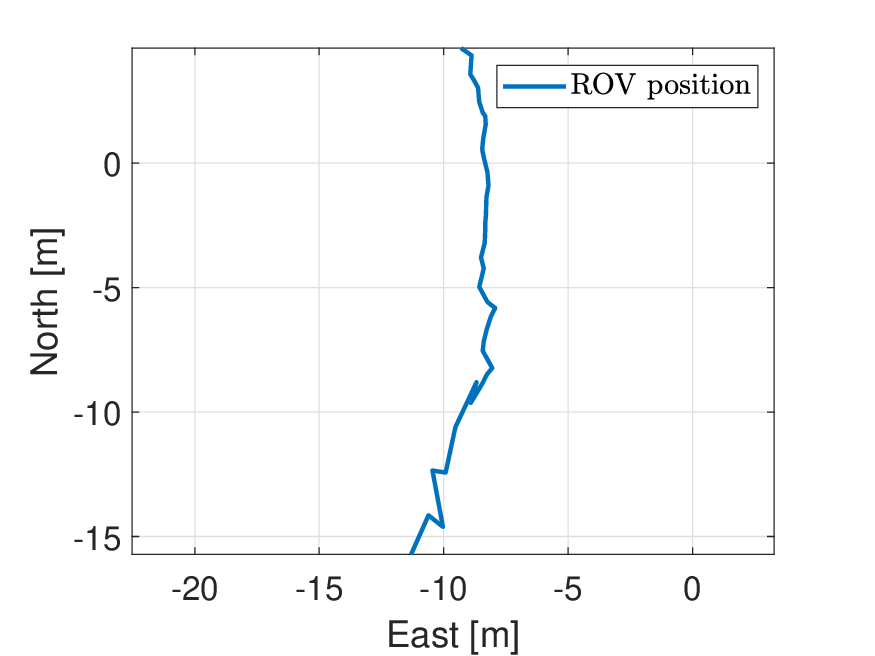}
    \caption{ROV position in the North-East frame.}
    \label{fig:rov_position}
\end{figure}

\begin{figure}
    \centering
    \includegraphics[width=\columnwidth]{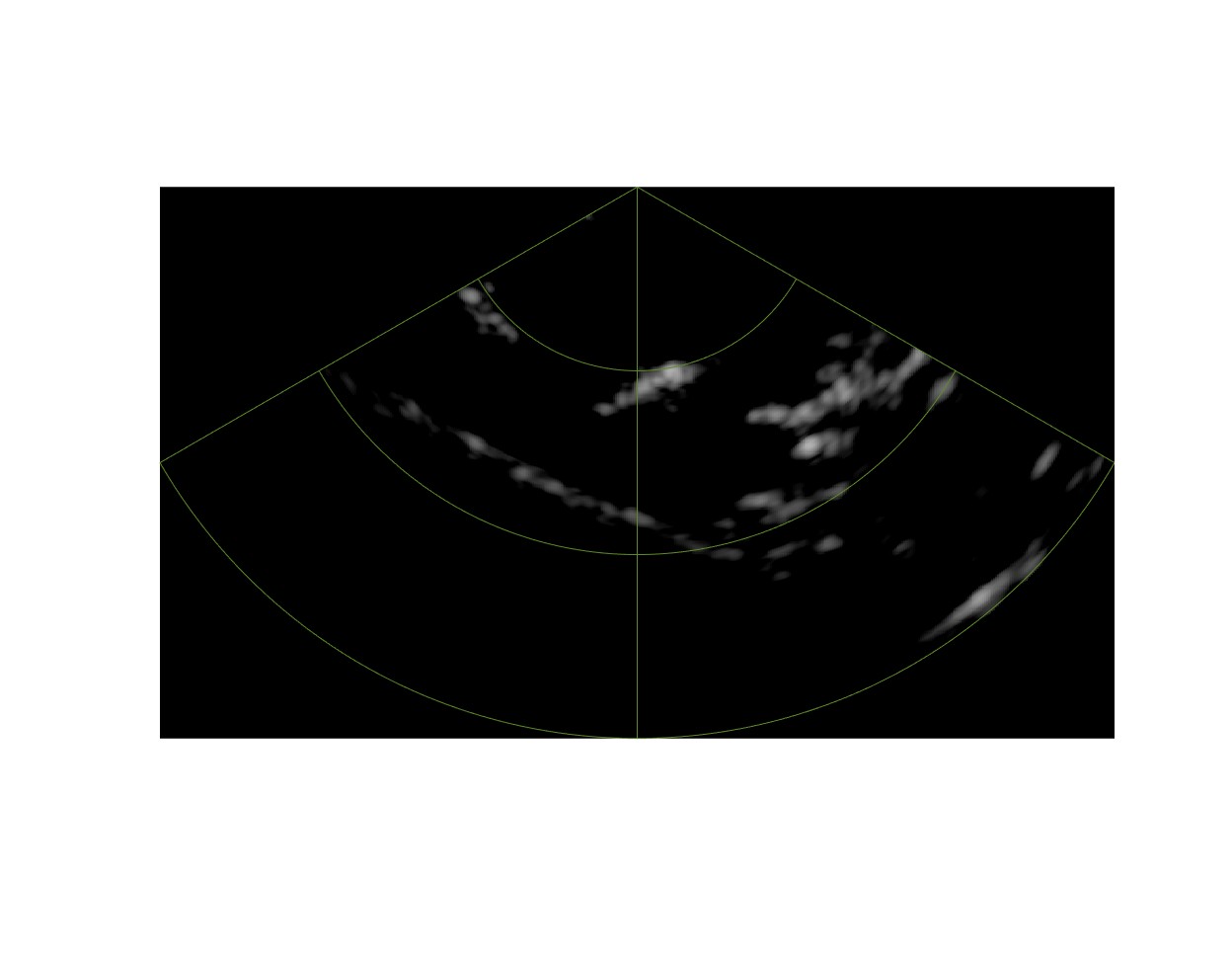}
    \vspace{-2cm}
    \caption{Image from the Sonoptix Multibeam sonar.}
    \label{fig:sonar}
\end{figure}

\begin{figure}
    \centering
    \includegraphics[width=\columnwidth]{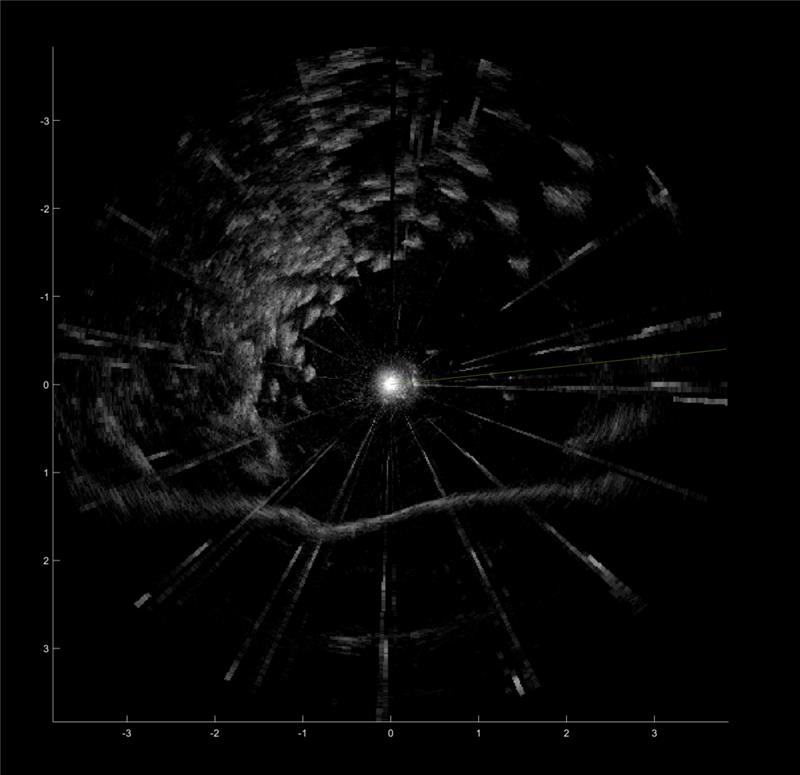}
    \caption{The data from the Ping360 sensor.}
    \label{fig:ping360}
\end{figure}

\begin{figure}
    \centering
    \includegraphics[width=\columnwidth]{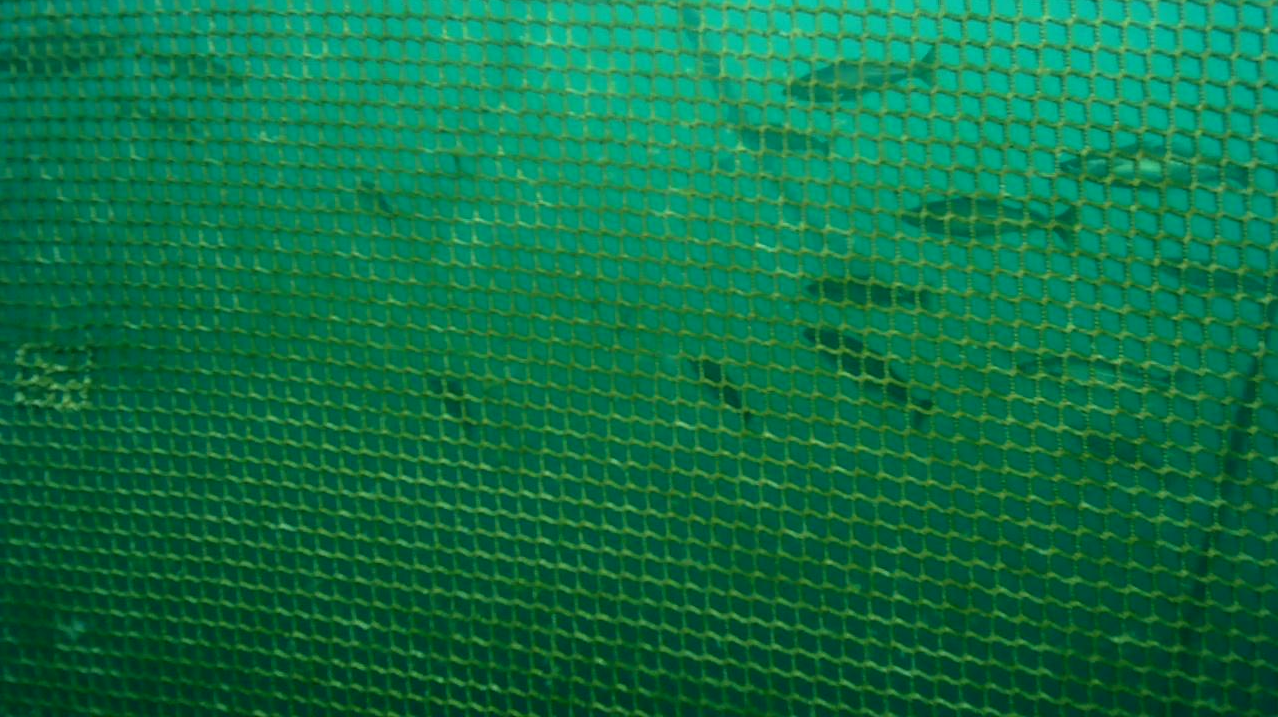}
    \caption{Screenshot from the mono camera during a trial.}
    \label{fig:mono_camera}
\end{figure}

\begin{figure}
    \centering
    \includegraphics[width=\columnwidth]{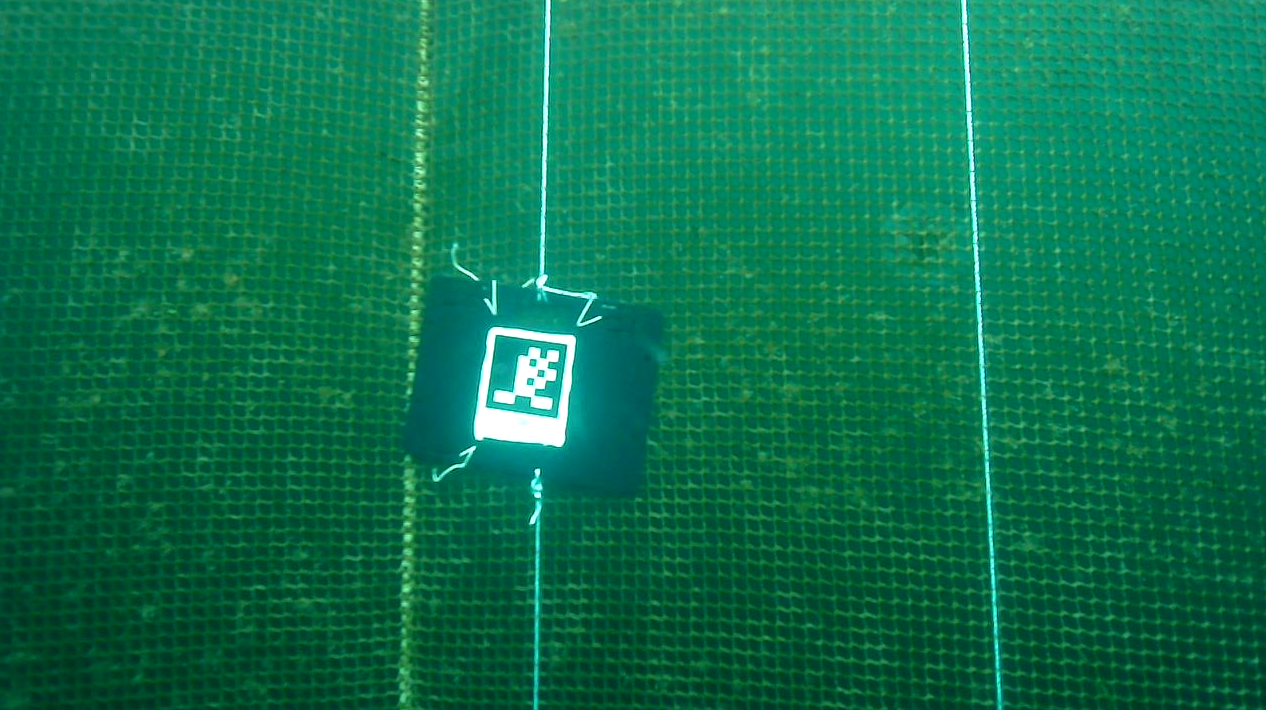}
    \caption{Screenshot from the mono camera showing the AprilTag}
    \label{fig:mono_camera_april_tag}
\end{figure}

\begin{figure}
    \centering
    \includegraphics[width=\columnwidth]{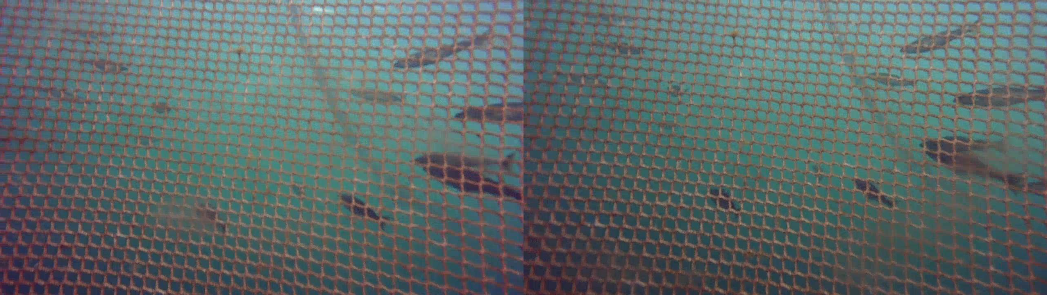}
    \caption{Screenshot from the stereo camera during a trial.}
    \label{fig:stereo_camera}
\end{figure}

In this section, examples of some of the collected data from the datasets will be presented (Figures~\ref{fig:both_DVLs} - Figure~\ref{fig:stereo_camera}). Figure~\ref{fig:both_DVLs} shows data from the Water Linked A50 DVL, the Nortek Nucleus 1000 DVL, and the desired speed of the ROV during one particular trial. Note that any deviation from the desired speed is not an indicator of sensor performance. Both sensors experience instances of invalid data points. These points are replaced by the latest valid data point for plotting purposes.

In addition to the raw measurements, using the DVL beam measurements from Water Linked A50 DVL and adapting the method developed in~\cite{amundsen2021autonomous}, the dataset includes estimations of the net-relative distance, net-relative heading and net-relative speed for each performed trials. Figures~\ref{fig:net_distance}, Figure ~\ref{fig:net_heading} and Figure~\ref{fig:net_velocity} show the net-relative distance, net-relative heading and net-relative speed for another trial, respectively.

The global position of the ROV is obtained using the Sonardyne MicroRanger 2 USBL during these trials. Figure~\ref{fig:rov_position} shows the ROV position in the North-East frame as captured by this system. From this it is possible to see that the ROV is following the curved shape of the net pen structure.

Although the sonar images are best viewed as a video stream, Figure~\ref{fig:sonar} shows one image from the Sonoptix Multibeam sonar containing the net structure and some fish. Figure~\ref{fig:ping360} shows a screenshot of data from the Ping360 sensor. The net and some fish are present. Lastly, Figure~\ref{fig:mono_camera}, Figure~\ref{fig:mono_camera_april_tag}, and Figure~\ref{fig:stereo_camera} show screenshots from the mono and stereo cameras, respectively.

\FloatBarrier

\section{Conclusions}
In this paper, we have presented a new large dataset, SOLAQUA, acquired during both manual and autonomous net following ROV operations in industrial scale fish farms in Norway. The proposed dataset contains data from the most commonly used acoustic and vision sensors in underwater domain. The released dataset covers a diverse set of software-synchronized measurements, including undamaged nets with some fish and marine growth presence. The datasets consists of numerous subsets from ROV operating in different depths with various speeds, keeping different net-relative distances from the net pen, and facing diverse environmental conditions. Our dataset will not only contribute to facilitate research on resilient underwater perception and robust navigation autonomy of underwater robotic systems operating in dynamic environments, but also provides large set of data from recording of net pen, relevant to develop and test methods for optimal and efficient inspection operations in net pens. This means that it is expected that both the research community and the aquaculture industry will benefit greatly from the utilization of the SOLAQUA dataset.

\section{Acknowledgements}
The authors express their sincerest gratitude to Kay Arne Skarpenes and Terje Bremvåg who assisted during the field trials. The work was funded by the Research Council of Norway through projects Resifarm (no. 327292) and CHANGE (no. 313737), and by SINTEF Ocean.

\bibliographystyle{ieeetr}
\bibliography{ref}

\end{document}